\pdfoutput=1
\documentclass[11pt]{article}

\usepackage[preprint]{acl}

\usepackage[T1]{fontenc}
\usepackage[utf8]{inputenc}
\usepackage{times}
\usepackage{latexsym}
\usepackage{microtype}
\usepackage{inconsolata}
\usepackage{xspace}

\usepackage{amsmath}
\usepackage{amssymb}
\usepackage{amsthm}
\usepackage{bm}
\usepackage{mathtools}

\usepackage{graphicx}
\usepackage{booktabs}
\usepackage{multirow}
\usepackage{subcaption}
\usepackage{makecell}
\usepackage{placeins}
\usepackage{needspace}

\usepackage{algorithm}
\usepackage{algpseudocode}

\usepackage{tikz}
\usetikzlibrary{
    arrows.meta,
    positioning,
    shapes.geometric,
    calc,
    fit,
    backgrounds,
    decorations.pathreplacing
}
\usepackage{pgfplots}
\pgfplotsset{compat=1.18}
\usepgfplotslibrary{groupplots, fillbetween}

\usepackage[capitalize,noabbrev]{cleveref}

\newcommand{\method}{HiMPO\xspace}
\newcommand{\mempo}{MemPO\xspace}

\newcommand{\code}[1]{\nolinkurl{#1}}

\newcommand{\zstar}{z^{\star}}

\newcommand{\Aoff}{\widetilde{A}^{M,\mathrm{eval}}}
\newcommand{\policy}{\pi_{\theta}}

\title{\method{}: Hindsight-Informed Memory Policy Optimization for Less-Entangled Credit in Long-Horizon Agents}

\author{
  \bfseries Jiangze Yan\textsuperscript{1,2} \quad Yi Shen\textsuperscript{1,2} \quad Wenjing Zhang\textsuperscript{1,2} \quad Jieyun Huang\textsuperscript{1,2} \quad Ping Chen\textsuperscript{1,2} \\
  \bfseries Zhaoxiang Liu\textsuperscript{1,2} \quad Ning Wang\textsuperscript{1,2} \quad Kai Wang\textsuperscript{1,2} \quad Shiguo Lian\textsuperscript{1,2} \\[4pt]
  \textsuperscript{1}Unicom Data Intelligence, China Unicom \\
  \textsuperscript{2}Data Science \& Artificial Intelligence Research Institute, China Unicom \\[2pt]
  \texttt{\{yanjz17, sheny73, liansg\}@chinaunicom.cn}
}

\hypersetup{
  pdfauthor={Jiangze Yan, Yi Shen, Wenjing Zhang, Jieyun Huang, Ping Chen, Zhaoxiang Liu, Ning Wang, Kai Wang, Shiguo Lian},
  pdftitle={HiMPO: Hindsight-Informed Memory Policy Optimization for Less-Entangled Credit in Long-Horizon Agents}
}

\begin{document}
\maketitle

\begin{abstract}

Long-horizon agents compress interaction histories into memory, but final outcomes
entangle memory-writing credit with tool failures and reasoning errors.
We propose \method{}, a Hindsight-Informed Memory Policy Optimization framework
that supplements trajectory-level reinforcement learning with memory-specific credit.
For each update, \method{} compares the target answer's log-likelihood under
the previous and updated memories while holding the remaining pre-write context fixed.
A soft gate based on group-relative hindsight relevance then weights the local
utility advantage. Trajectory-level credit applies to all generated tokens;
memory tokens additionally receive the auxiliary advantage.
Across judge-based open-domain tasks and objective compressive-memory QA,
\method{} improves task performance while preserving compressed-context efficiency.
On Local Wiki Search, it improves average F1 and exact match over \mempo{}
by 2.6 and 2.2 percentage points, respectively.
Controlled interventions and live memory-reversion studies provide evidence of
reduced blame leakage and alignment between memory credit and behavioral effects.
With the policy fixed, the credit estimator remains robust to the tested
forms of target noise.

\end{abstract}

\section{Introduction}
\label{sec:intro}

Long-horizon agents solve complex tasks by interleaving reasoning, tool invocation, and environmental feedback, as in web search~\citep{wu2025resum}, multi-hop question answering~\citep{zhao2025reagent}, embodied planning~\citep{qian2025discriminator}, and tool-augmented reasoning~\citep{wu2025agenticreasoning}. However, full-history prompting causes the context to grow with the number of interaction steps, increasing token cost, stressing fixed context windows, and degrading performance under long contexts~\citep{liu2024lostmiddle}. Memory compression is therefore a key capability for efficient long-horizon agent learning.

\begin{figure}[!t]
\centering
\includegraphics[width=\columnwidth]{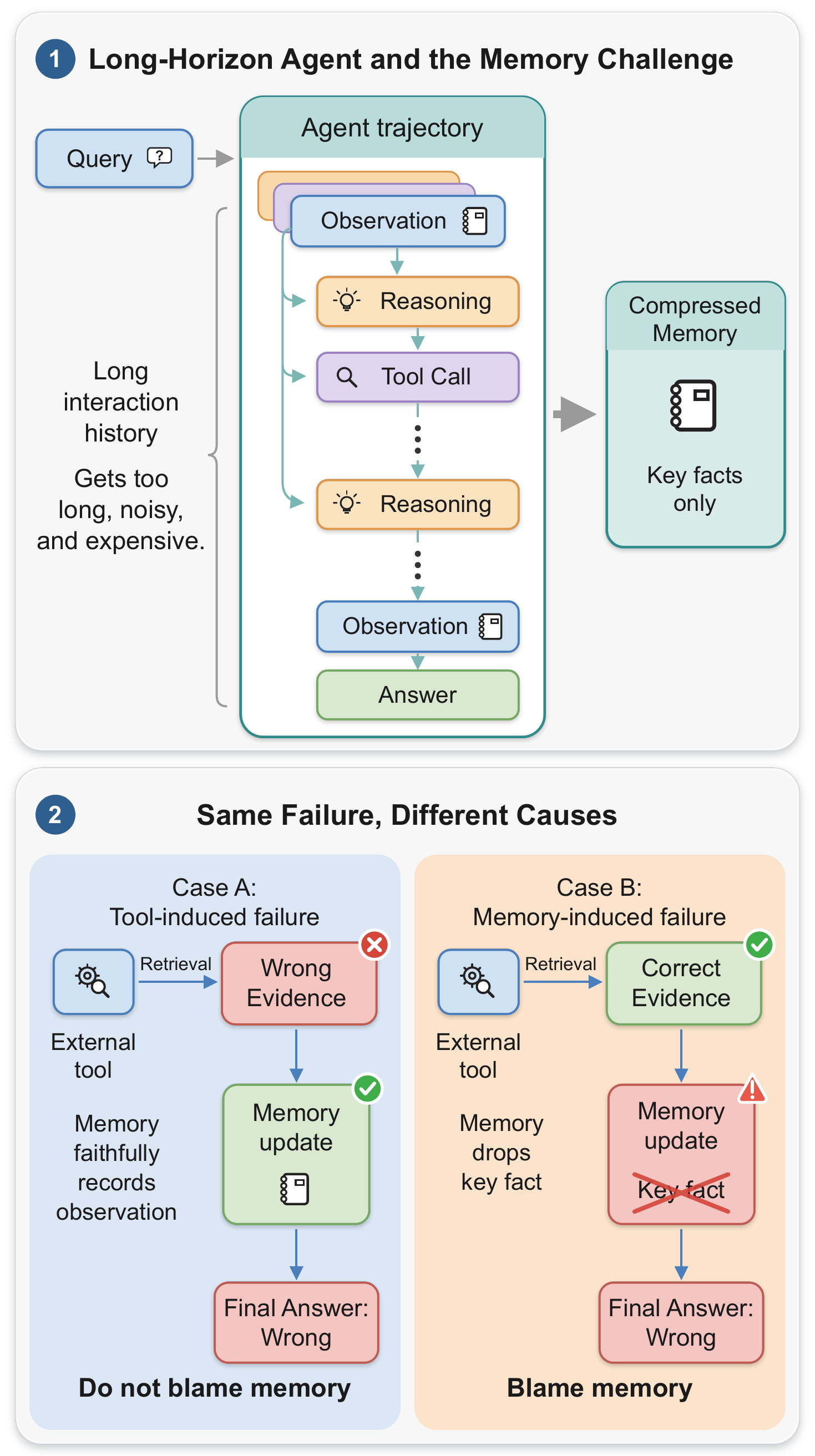}
\caption{Causally entangled memory credit. The same final failure can require opposite memory credits depending on whether the error originates from tools or memory. In Case A, the tool returns wrong evidence and a faithful memory should not be blamed; in Case B, the tool returns correct evidence but the memory drops a key fact and should receive negative credit.}
\label{fig:intro}
\end{figure}

Existing memory mechanisms retrieve stored interactions on demand,
compress histories into compact agent states, or, most directly, integrate
memory generation into policy learning; retrieval-based access, however,
may be weakly aligned with the task objective.
End-to-end compression enables task-aware memory writing, but it also
introduces a new credit assignment problem: final outcomes are jointly
determined by memory updates, tool calls, observations, and reasoning,
making it unclear which component should be rewarded or penalized.

For end-to-end memory compression, the credit assigned to memory updates is
often causally entangled with other agent components. As illustrated in
\Cref{fig:intro}, the same failed trajectory may require opposite
memory credits. In Case A, the tool retrieves incorrect evidence and the
memory faithfully summarizes it; the final answer is wrong, but the memory
itself should not be penalized. In Case B, the tool retrieves correct
evidence but the memory drops a key fact, so the memory should receive
negative credit. Outcome-based memory rewards cannot distinguish these
cases because both lead to failure---a problem we call \emph{causally
entangled memory credit assignment}: tool-, observation-, or
reasoning-induced errors are incorrectly propagated to memory updates.

To reduce this entanglement, we argue that memory credit should depend on
two signals. \emph{Local counterfactual utility} asks
whether, under the same compressed pre-write state, replacing the
previous memory with the updated one improves the recoverability of
task-relevant information; \emph{hindsight relevance} asks whether,
conditioned on the target outcome, the update becomes more likely
relative to sibling memory writes. \method{} operationalizes this idea by treating each memory
update as a policy-controlled writing action, estimating its local utility
through a memory-replacement counterfactual, and using hindsight relevance
through a retrospective filter that weights local utility. The filtered
memory-specific advantage is applied only to \texttt{<mem>} tokens, while
trajectory-level rewards provide global training stability.

We evaluate \method{} under two evaluation protocols---judge-based
open-domain long-horizon tasks and objective compressive-memory
QA---where it improves over strong RL-based and memory-based baselines
under compressed-context inference; controlled interventions further show
that it reduces blame leakage from tool-induced errors and better
localizes credit to the affected memory updates.

Our main contributions are summarized as follows:
\begin{itemize}
\item We identify causally entangled memory credit assignment as a key obstacle in training long-horizon agents with compressed memory.

\item We propose \method{}, which uses a local memory-state counterfactual as the primary credit signal and a sign-consistent retrospective filter to weight that signal by hindsight relevance, enabling more faithful optimization of memory writing.

\item We conduct extensive experiments and controlled analyses on long-horizon agent benchmarks, showing that \method{} improves task performance under compressed-context inference and reduces blame leakage from non-memory errors.

\end{itemize}

\section{Related Work}
\label{sec:related}

\paragraph{Memory management for LLM agents.}
Memory mechanisms have been widely studied to mitigate context growth in long-horizon LLM agents. External-memory systems such as MemGPT~\citep{packer2023memgpt}, MemoryBank~\citep{zhong2024memorybank}, Mem0~\citep{chhikara2025mem0}, and A-Mem~\citep{a_mem_2024} store past interactions and retrieve relevant fragments when needed, but their memory access---self-directed paging in MemGPT, similarity-based retrieval in the others---is often weakly aligned with task-level objectives. Summarization-based methods such as ReSum~\citep{wu2025resum} instead compress accumulated histories into compact reasoning states, but do not determine whether a specific memory update should be credited for the final outcome.

\paragraph{End-to-end memory compression and self-memory optimization.}
Recent methods integrate memory or summary generation into agent training. SUPO~\citep{supo2025} incorporates summarization-based context management into multi-turn RL, MEM1~\citep{mem1_2024} studies memory-reasoning synergy, and MemPO~\citep{mempo2024} introduces a \texttt{<mem>} action optimized with trajectory- and memory-level advantages. These methods enable task-aware compression, but their supervision remains largely outcome-driven: memory updates are rewarded by final success or answer likelihood, without distinguishing errors caused by memory, tools, observations, or reasoning. Unlike prior self-memory optimization methods, \method{} treats each memory update as a state-writing transition and assigns credit by comparing the updated memory against the previous memory under the same pre-write state, shifting the optimization target from scoring a compressed memory state to estimating the incremental utility of a memory write.

\paragraph{Credit assignment for long-horizon agent optimization.}
Credit assignment is challenging in long-horizon agent optimization, where sparse outcome rewards provide coarse supervision. Value-free RL methods such as GRPO~\citep{grpo2024} estimate advantages from group rewards, but often fail to identify pivotal intermediate decisions. Finer-grained methods use entropy-based shaping, process supervision, state grouping, or hindsight signals~\citep{harutyunyan2019hca}, including GTPO/GRPO-S~\citep{tgrpo2025}, GiGPO~\citep{feng2025gigpo}, EMPG~\citep{wang2025empg}, and HCAPO~\citep{hcapo2024}; process reward models and step-level verification~\citep{uesato2022process,lightman2023verify,wang2024mathshepherd} likewise densify supervision at the step level. However, these methods target general actions or reasoning steps; finer temporal granularity alone cannot distinguish ``wrong tool, faithful memory'' from ``correct tool, harmful memory''. \method{} instead takes the memory transition $m_{t-1}\!\rightarrow\!m_{t}$, evaluated under an identical pre-write state, as the unit of credit, combining local counterfactual utility with a retrospective filter.

\section{Method}
\label{sec:method}

\begin{figure*}[!t]
\centering
\includegraphics[width=\textwidth]{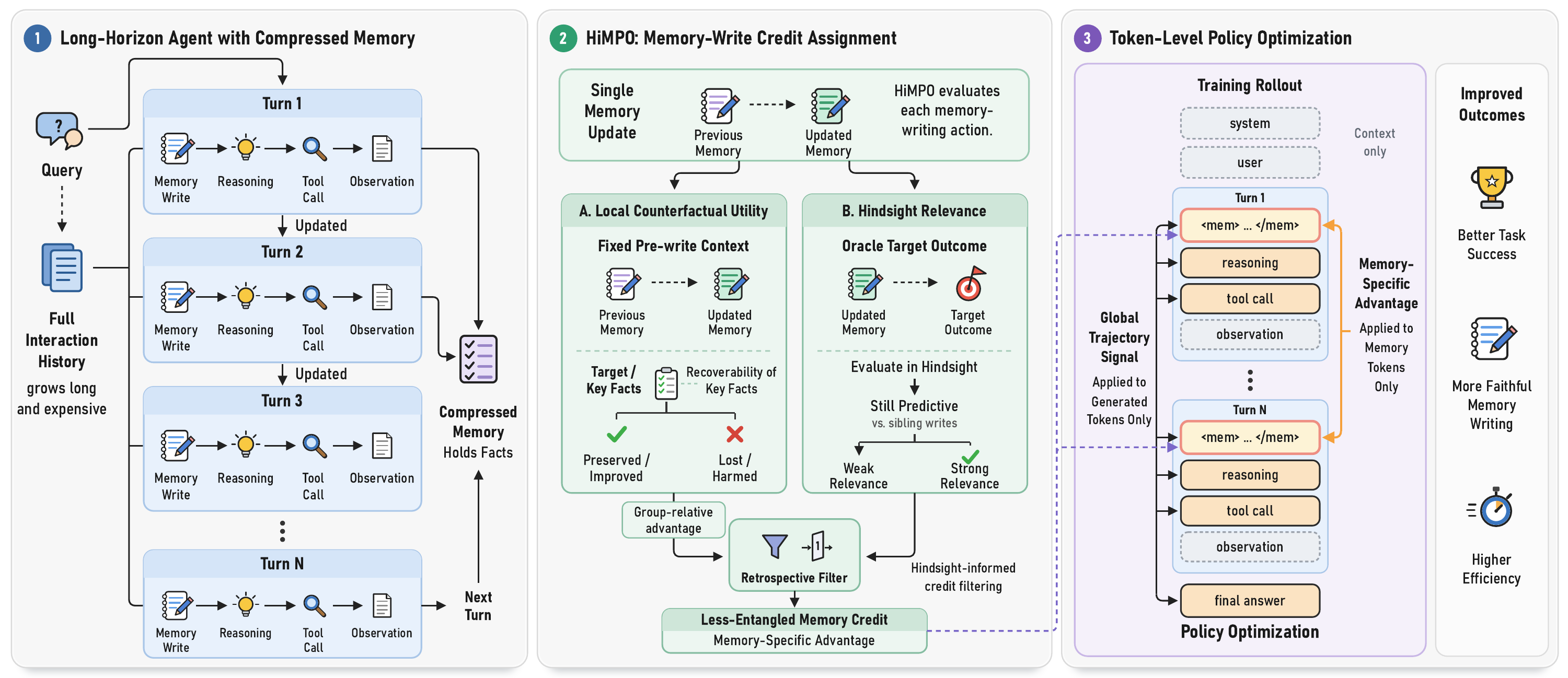}
\caption{Overview of \method{}. Local utility from memory replacement is standardized within each group and step, then weighted by a retrospective filter. All policy-generated tokens receive trajectory-level credit; \texttt{<mem>} tokens additionally receive the memory-specific advantage. System/user text and tool observations provide context only.}
\label{fig:method}
\end{figure*}

\paragraph{Method at a glance.}
\Cref{fig:method} summarizes memory-credit computation. During training,
the rollout policy performs three teacher-forced scoring passes per write:
(i) the target answer under the updated memory; (ii) the same answer under
the previous memory, with the remaining pre-write context fixed; and
(iii) the generated memory conditioned on the target answer.
All three scores come from the rollout policy itself; no external
reward model or LLM judge is used during training.
The difference between (i) and (ii) gives local counterfactual utility
(\Cref{sec:local_utility}); (iii) gives hindsight relevance
(\Cref{sec:hindsight_gate}). The filter combines these
signals, giving larger weights when their group-relative signs agree.
The resulting memory advantage supplements trajectory-level credit on
\texttt{<mem>} tokens (\Cref{sec:token_opt}). The scoring passes use
existing tokens without additional autoregressive decoding.

\subsection{Problem Setup and Overview}
\label{sec:setup_overview}

We consider long-horizon agent tasks where an LLM-based policy interacts
with an external environment through multi-turn reasoning and tool use.
Given a task query $q_i$, the agent maintains a compressed memory state
across interaction rounds. At step $t$, the policy first writes a memory
summary $m_{i,t}$, then produces a reasoning segment $r_{i,t}$ and a
tool call $c_{i,t}$; the environment subsequently returns an observation
$o_{i,t}$ according to the issued tool call. We denote the pre-write
state before generating the current memory as
\begin{equation}
H_{i,t} = (q_i, m_{i,t-1}, r_{i,t-1}, c_{i,t-1}, o_{i,t-1}),
\end{equation}
where $m_{i,t-1}$ is the previous compressed memory and
$(r_{i,t-1}, c_{i,t-1}, o_{i,t-1})$ is the most recent interaction.
For the first step, we use an empty memory $m_{\emptyset}$ and omit the
previous interaction. The memory write at step $t$ can therefore be
viewed as a policy-controlled action
$m_{i,t} \sim \pi_\theta(\cdot \mid H_{i,t})$, which updates the agent
state from $m_{i,t-1}$ to $m_{i,t}$.

A complete rollout is denoted as
$\tau_i = \{(m_{i,t}, r_{i,t}, c_{i,t}, o_{i,t})\}_{t=0}^{T_i}$,
where $T_i$ is the index of the final interaction (and memory-writing)
step; the rollout ends with a final answer or terminal decision. Each trajectory
receives a rule-based outcome reward $R^T(\tau_i)$: exact match of the
extracted final answer against the dataset gold answer, gated by a
format-validity check that zeroes the reward of malformed trajectories.
Following group-based policy
optimization~\citep{grpo2024}, for a group of $N$ rollouts sampled from the same query,
we compute the trajectory-level advantage as
\begin{equation}
A_i^T =
\frac{R^T(\tau_i)-\mu_R}{\sigma_R+\epsilon},
\end{equation}
where $\mu_R$ and $\sigma_R$ are the mean and standard deviation of
trajectory rewards within the group, and $\epsilon$ is a small constant
for numerical stability (also used in
\Cref{eq:delta_norm,eq:beta_eff}). This trajectory-level signal
provides coarse global supervision for all policy-generated tokens.

However, a final success or failure is jointly affected by memory
updates, tool calls, observations, and subsequent reasoning, so
directly propagating the outcome reward to memory tokens leads to
causally entangled credit. \method{} therefore assigns each memory write
$m_{i,t}$ a memory-specific advantage
$A^M_{i,t}=G(\widehat{\Delta}_{i,t}, \log\rho_{i,t})\cdot
\widehat{\Delta}_{i,t}$ (derived in \Cref{sec:hindsight_gate}),
where $\widehat{\Delta}_{i,t}$ is the group-relative advantage obtained
by standardizing the update's raw local utility $\Delta_{i,t}$,
$\log\rho_{i,t}$ measures its hindsight relevance with respect to the
target outcome $z_i^\star$, and the filter $G(\cdot)$ combines
the two signals to weight memory credit. All policy-generated tokens
receive $A_i^T$; only \texttt{<mem>} tokens additionally receive the
memory-specific advantage.

\subsection{Local Counterfactual Utility}
\label{sec:local_utility}

The raw local utility measures the change in target-answer recoverability
caused by replacing the previous memory. Let $H_{i,t}[m]$ denote the
pre-write state with its memory slot replaced by candidate $m$, keeping
the query and recent interaction fixed. The answerability score is the
average log-likelihood of the target outcome:
\begin{equation}
\label{eq:answerability}
\resizebox{0.88\columnwidth}{!}{$\displaystyle
S(H_{i,t}, m, z_i^\star)
=
\frac{1}{|z_i^\star|}
\sum_{\ell=1}^{|z_i^\star|}
\log \pi_{\theta_{\mathrm{old}}}\!\left(
z_{i,\ell}^\star \mid H_{i,t}[m], z_{i,<\ell}^\star
\right).
$}
\end{equation}
Here $z_{i,\ell}^\star$ denotes the $\ell$-th token of the target
outcome $z_i^\star$, $|z_i^\star|$ its token length, and
$\pi_{\theta_{\mathrm{old}}}$ the rollout policy. Auxiliary scores are
computed before optimization and held fixed during each batch of updates.

We define the raw local counterfactual utility of the memory update
$m_{i,t-1}\!\rightarrow m_{i,t}$ as
\begin{equation}
\resizebox{0.89\columnwidth}{!}{$\displaystyle
\label{eq:delta}
\Delta_{i,t}
=
S(H_{i,t}, m_{i,t}, z_i^\star)
-
S(H_{i,t}, m_{i,t-1}, z_i^\star).
$}
\end{equation}
By comparing the updated memory with the previous memory under the same
pre-write state, $\Delta_{i,t}$ estimates the marginal utility of the
current memory write rather than the cumulative quality of the whole
memory history. For the first memory write, we use the empty memory
$m_{\emptyset}$ as the counterfactual baseline, i.e.,
$\Delta_{i,0}=S(H_{i,0}, m_{i,0}, z_i^\star)-S(H_{i,0}, m_{\emptyset},
z_i^\star)$.
This comparison also matches Case A in \Cref{fig:intro}: because the
most recent observation $o_{i,t-1}$ remains in $H_{i,t}$ under both
memories, a write that faithfully restates a misleading observation adds
little answerability beyond the observation itself, so its local utility
stays near zero instead of inheriting the trajectory's failure.

Since answerability scores may vary across interaction steps, we
standardize $\Delta_{i,t}$ within each rollout group and step index:
\begin{equation}
\label{eq:delta_norm}
\widehat{\Delta}_{i,t}
=
\frac{
\Delta_{i,t}-\mu_{\Delta}^{(g,t)}
}{
\sigma_{\Delta}^{(g,t)}+\epsilon
},
\end{equation}
where $\mu_{\Delta}^{(g,t)}$ and $\sigma_{\Delta}^{(g,t)}$ are computed
over rollouts from the same prompt group $g$ that contain a memory write
at step $t$. The resulting group-relative utility advantage
$\widehat{\Delta}_{i,t}$ measures utility relative to this group
baseline: its sign need not match that of the raw utility $\Delta_{i,t}$.

\subsection{Retrospective Filtering of Memory Utility}
\label{sec:hindsight_gate}

Hindsight relevance modulates \(\widehat{\Delta}_{i,t}\) through the filter;
it is not a standalone reward. We view each memory update as a writing action
$w_{i,t}: m_{i,t-1}\!\rightarrow m_{i,t}$. Given the pre-write state
$H_{i,t}$ and the target outcome $z_i^\star$, we compute the hindsight
likelihood of the generated memory tokens $y_{i,t,1:|m_{i,t}|}$:
\begin{equation}
\resizebox{0.89\columnwidth}{!}{$\displaystyle
\label{eq:hindsight_likelihood}
\log h_{i,t}
=
\frac{1}{|m_{i,t}|}
\sum_{j=1}^{|m_{i,t}|}
\log \pi_{\theta_{\mathrm{old}}}
\left(
y_{i,t,j}
\mid
H_{i,t}, z_i^\star, y_{i,t,<j}
\right).
$}
\end{equation}
Here, $y_{i,t,j}$ is the $j$-th token of $m_{i,t}$, and the context is
formed by appending the oracle target outcome $z_i^\star$ (not the
agent's generated answer) to $H_{i,t}$ under a fixed template
(\Cref{app:hindsight_template}). Intuitively, if a memory write is
important for reaching the target outcome, it should become more likely
when the policy is conditioned on that outcome.

Since directly estimating the prior probability of a natural-language
memory write is difficult, we use a self-normalized proxy of the
hindsight likelihood ratio. For each prompt group and step index, we
center the hindsight likelihood by sibling rollouts:
\begin{equation}
\label{eq:rho}
\log \rho_{i,t}
=
\log h_{i,t}
-
\frac{1}{|\mathcal{G}_{i,t}|}
\sum_{j\in\mathcal{G}_{i,t}}
\log h_{j,t},
\end{equation}
where $\mathcal{G}_{i,t}$ contains rollouts from the same prompt group
that produce a memory write at step $t$; $\log\rho_{i,t}>0$ thus
indicates that the write is more hindsight-relevant than its siblings.

We combine hindsight relevance with group-relative utility using a
sign-consistent retrospective filter:
\begin{equation}
\resizebox{0.89\columnwidth}{!}{$\displaystyle
\label{eq:sign_gate}
G(\widehat{\Delta}_{i,t}, \log\rho_{i,t})
=
\sigma\left(
\beta_{\mathrm{eff}}
\left(
\operatorname{sgn}(\widehat{\Delta}_{i,t})\log\rho_{i,t}
-
\tau_\rho
\right)
\right),
$}
\end{equation}
where $\sigma(\cdot)$ is the sigmoid function, $\tau_\rho$ is a filter
threshold, and $\beta_{\mathrm{eff}}$ controls the sharpness of the
filter. We write $G_{i,t}:=G(\widehat{\Delta}_{i,t},\log\rho_{i,t})$
for the scalar filter weight. The resulting memory-specific advantage is
\begin{equation}
\label{eq:memory_adv}
A^M_{i,t}
=
G(\widehat{\Delta}_{i,t}, \log\rho_{i,t})
\cdot
\widehat{\Delta}_{i,t}.
\end{equation}

The sigmoid assigns larger weights when the group-relative utility
advantage $\widehat{\Delta}_{i,t}$ and centered hindsight score
$\log\rho_{i,t}$ agree in sign. Otherwise, it attenuates the credit
without setting it to zero.

\subsection{Stabilized Token-Level Policy Optimization}
\label{sec:token_opt}

Directly applying the filtered advantage in \Cref{eq:memory_adv} to all
memory updates may introduce instability, so we adopt two stabilization
mechanisms before scattering it to tokens.

First, a memory-success mask suppresses negative auxiliary memory
advantages on trajectories that reach the target outcome:
\begin{equation}
\label{eq:do_no_harm}
A^M_{i,t} \leftarrow 0
\quad
\text{if }
R^T(\tau_i) \geq \tau_{\mathrm{succ}}
\ \text{and}\
A^M_{i,t}<0.
\end{equation}

Second, since some memory updates are prefatory and only become useful
after later observations are collected, we optionally propagate memory
credit backward along the memory sequence using an exponential moving
average:
\begin{equation}
\label{eq:smoothing}
\widetilde{A}^M_{i,t}
=
\alpha A^M_{i,t}
+
(1-\alpha)\widetilde{A}^M_{i,t+1},
\end{equation}
with the boundary condition $\widetilde{A}^M_{i,T_i}=A^M_{i,T_i}$ at
the final memory-writing step $T_i$, and $\alpha$ the smoothing
coefficient. Here $A^M_{i,t}$ is the filtered credit after the
memory-success mask, while $\widetilde{A}^M_{i,t}$ is the final
memory credit used for training. If smoothing is disabled, we set
$\widetilde{A}^M_{i,t}=A^M_{i,t}$. Backward smoothing allows downstream
memory utility to flow back to earlier prerequisite writes.

Let $\mathcal{O}_i$ index the policy-generated tokens in $\tau_i$;
system/user text and tool observations are excluded from the loss.
For $k\in\mathcal{O}_i$, the token-level advantage is
\begin{equation}
\resizebox{0.88\columnwidth}{!}{$\displaystyle
\label{eq:scatter}
A_{i,k}
=
\begin{cases}
A_i^T + \lambda_M \widetilde{A}^M_{i,t(k)},
& k \in \mathrm{span}(m_{i,t(k)}),\\[2pt]
A_i^T,
& \text{otherwise},
\end{cases}
$}
\end{equation}
where $t(k)$ denotes the memory step associated with token $k$, and
$\lambda_M$ controls the strength of the memory-specific signal. The
policy is then optimized with the standard clipped policy
objective~\citep{schulman2017ppo}:
\begin{equation}
\label{eq:ppo_obj}
\begin{aligned}
&\mathcal{J}(\theta)
=\mathbb{E}\Big[
\frac{1}{|\mathcal{O}_i|}
\sum\nolimits_{k\in\mathcal{O}_i}
\Big\{\min\Big(
r_{i,k}(\theta)A_{i,k}, \\
&\quad\mathrm{clip}(r_{i,k}(\theta),1-\epsilon_{\mathrm{clip}},1+\epsilon_{\mathrm{clip}})A_{i,k}
\Big) \\
&\quad-\beta_{\mathrm{KL}}
D_{i,k}(\theta)\Big\}
\Big],
\end{aligned}
\end{equation}
where
$r_{i,k}(\theta)=
\pi_\theta(a_{i,k}\mid x_{i,k})/
\pi_{\theta_{\mathrm{old}}}(a_{i,k}\mid x_{i,k})$.
Here $a_{i,k}$ is the $k$-th policy-generated token (token-level
action), $\pi_{\theta_{\mathrm{old}}}$ is the rollout policy, and $x_{i,k}$
is the rollout context for token $k$,
including retained history and the generated prefix within the turn;
both policies use the same context. The KL term is
$D_{i,k}(\theta)=D_{\mathrm{KL}}(\pi_\theta(\cdot\mid x_{i,k})\|
\pi_{\mathrm{ref}}(\cdot\mid x_{i,k}))$, where $\pi_{\mathrm{ref}}$
is a frozen copy of the initial policy checkpoint. We set $\beta_{\mathrm{KL}}=0.001$ (\Cref{tab:hparams});
$\epsilon_{\mathrm{clip}}$ is the PPO clipping range.
Thus, all generated tokens receive global credit, with additional
memory-specific credit only for memory tokens.

\subsection{Training Procedure}
\label{sec:algorithm}

For each rollout group, \method{} computes trajectory advantages from
terminal rewards and memory advantages from local utility and the
retrospective filter. The memory-success mask and backward smoothing
stabilize the memory advantages before they are assigned to
\texttt{<mem>} tokens. The policy is then updated with the objective in
\Cref{eq:ppo_obj}. At inference time, it maintains compressed memory
using the same \texttt{<mem>}-based interaction protocol.
\Cref{alg:himpo} gives the training algorithm.

\begin{table*}[!t]
\centering
\small
\setlength{\tabcolsep}{6pt}
\renewcommand{\arraystretch}{1.15}
\begin{tabular*}{\textwidth}{@{\extracolsep{\fill}}l l cc cc cc c@{}}
\toprule
& & \multicolumn{7}{c}{Accuracy (\%, $\uparrow$)} \\
\cmidrule(lr){3-9}
Benchmark & Method
& \multicolumn{2}{c}{4k ctx.}
& \multicolumn{2}{c}{8k ctx.}
& \multicolumn{2}{c}{16k ctx.}
& full ctx. \\
\cmidrule(lr){3-4} \cmidrule(lr){5-6} \cmidrule(lr){7-8}
& & M & R & M & R & M & R & \\
\midrule

\multirow{5}{*}{BrowseComp-Plus}
& ReAct
& 7.47 & 5.18
& 6.27 & 5.30
& 5.30 & 6.39
& 5.06 \\
& SFT-only
& 7.43 & 2.45
& 7.79 & 7.39
& 6.91 & 6.71
& 5.86 \\
& \mempo{}
& 7.75 & 3.05
& 9.40 & 8.71
& 7.71 & 7.47
& 7.03 \\
& SUPO
& 8.47 & 3.25
& 9.60 & 9.48
& 8.51 & 7.95
& 7.83 \\
& \textbf{\method{} (Ours)}
& \textbf{11.08} & \textbf{6.14}
& \textbf{12.17} & \textbf{11.65}
& \textbf{10.72} & \textbf{10.28}
& \textbf{10.84} \\

\midrule

\multirow{5}{*}{FRAMES}
& ReAct
& 17.84 & 17.23
& 16.02 & 20.63
& 18.57 & 16.87
& 17.11 \\
& SFT-only
& 21.16 & 21.84
& 21.36 & 21.93
& 21.84 & 22.45
& 21.36 \\
& \mempo{}
& 20.87 & 21.16
& 21.12 & 21.40
& 20.63 & 20.11
& 21.88 \\
& SUPO
& 22.13 & 22.86
& 21.80 & 21.88
& 21.52 & 21.80
& 22.17 \\
& \textbf{\method{} (Ours)}
& \textbf{24.96} & \textbf{25.16}
& \textbf{24.03} & \textbf{26.01}
& \textbf{25.20} & \textbf{23.79}
& \textbf{25.12} \\

\bottomrule
\end{tabular*}
\par
\caption{Main results on judge-based open-domain tasks. M and R denote the \texttt{mem\_aware} and \texttt{naive\_recency} truncation strategies (\Cref{app:truncation}). All trainable rows are reproduced from the same in-house Qwen2.5-7B-Instruct SFT initialization with a matched evaluation stack.}
\label{tab:budget_sweep}
\end{table*}

\begin{table*}[!t]
\centering
\footnotesize
\setlength{\tabcolsep}{3pt}
\renewcommand{\arraystretch}{1.15}
\begin{tabular*}{\textwidth}{@{\extracolsep{\fill}}l cc cc cc cc cc cc cc@{}}
\toprule
                  & \multicolumn{12}{c}{Local Wiki Search (F1 / EM, \%; $\uparrow$)}
                  & \multicolumn{2}{c}{Efficiency ($\downarrow$)} \\
\cmidrule(lr){2-13} \cmidrule(lr){14-15}
                  & \multicolumn{2}{c}{2-objective}
                  & \multicolumn{2}{c}{4-objective}
                  & \multicolumn{2}{c}{6-objective}
                  & \multicolumn{2}{c}{8-objective}
                  & \multicolumn{2}{c}{10-objective}
                  & \multicolumn{2}{c}{Average}
                  & \multicolumn{2}{c}{Average} \\
\cmidrule(lr){2-3} \cmidrule(lr){4-5} \cmidrule(lr){6-7}
\cmidrule(lr){8-9} \cmidrule(lr){10-11} \cmidrule(lr){12-13}
\cmidrule(lr){14-15}
Model             & F1 & EM & F1 & EM & F1 & EM & F1 & EM & F1 & EM & F1 & EM & TT & PT \\
\midrule
ReAct   & 33.7 & 25.6 & 10.6 &  7.0 &  5.4 &  4.0 &  5.9 &  4.3 &  2.6 &  2.0 & 11.7 &  8.6 & 3.6 & 0.6 \\
ReSearch          & 47.4 & 36.0 & 24.1 & 16.7 & 20.8 & 15.9 & 10.8 &  7.7 &  5.1 &  3.6 & 21.7 & 16.0 & 3.3 & 0.7 \\
DeepResearcher    & 30.9 & 24.7 & 24.5 & 18.1 & 13.9 & 10.7 &  9.1 &  6.9 &  5.1 &  3.5 & 16.7 & 12.8 & 4.3 & 0.8 \\
A-Mem             & 33.2 & 25.2 & 13.7 & 10.1 &  9.8 &  7.1 &  6.9 &  5.1 &  5.6 &  3.6 & 13.8 & 10.2 & 2.6 & 0.4 \\
MEM1              & 47.7 & 37.1 & 26.5 & 18.9 & 18.8 & 14.1 & 19.0 & 13.6 & 19.6 & 13.4 & 26.3 & 19.4 & 1.4 & \textbf{0.2} \\
GRPO (w/o mem)    & 54.6 & 43.0 & 38.3 & 28.6 & 29.8 & 22.6 & 19.0 & 13.7 & 11.0 &  7.8 & 30.5 & 23.1 & 4.4 & 0.8 \\
\mempo{}          & 56.5 & 46.2 & 42.8 & 31.9 & 34.3 & 26.9 & 30.5 & 23.7 & 24.2 & 18.2 & 37.6 & 29.4 & \textbf{1.2} & \textbf{0.2} \\
SUPO
                  & 57.9 & 47.2 & 43.0 & 31.8 & 30.3 & 24.0 & 22.7 & 17.5 & 16.3 & 11.6 & 34.0 & 26.4 & 1.4 & \textbf{0.2} \\
\midrule
\textbf{\method{} (Ours)}
                  & \textbf{61.3} & \textbf{51.0}
                  & \textbf{47.5} & \textbf{36.8}
                  & \textbf{35.1} & \textbf{27.2}
                  & \textbf{31.9} & \textbf{24.4}
                  & \textbf{25.1} & \textbf{18.6}
                  & \textbf{40.2} & \textbf{31.6}
                  & \textbf{1.2} & \textbf{0.2} \\
\bottomrule
\end{tabular*}
\caption{Results on multi-objective Local Wiki Search. \method{} and SUPO are initialized from the public \mempo{} SFT checkpoint and trained/evaluated under the original \mempo{} protocol. We verified stack compatibility by reproducing \mempo{} within $\pm 0.5$ F1/EM of the reported numbers, and therefore cite the remaining baseline rows directly from the \mempo{} paper~\citep{mempo2024}.}
\label{tab:multi_obj}
\end{table*}

\section{Experiments}
\label{sec:exp}
\subsection{Experimental Setup}
\label{sec:setup}

\paragraph{Benchmarks and metrics.}
We evaluate \method{} under two evaluation protocols. For judge-based open-domain long-horizon evaluation, we use BrowseComp-Plus~\citep[BCP;][]{browsecomp_plus} and FRAMES~\citep{frames2024}, where agents must retrieve, compress, and integrate evidence across multi-step interactions; answers in these benchmarks are judged by an LLM against the reference answer, so we report judge-based accuracy following the benchmark protocol. For objective compressive-memory QA, we use Local Wiki Search, the compressive-memory benchmark released with \mempo{}~\citep{mempo2024}, where gold answers enable word-level F1 and exact-match (EM) evaluation. $K$ denotes the number of objectives in each composed Local Wiki Search task. For BCP, we additionally conduct a context-budget sweep to assess memory retention under truncation pressure. We also report total-token consumption per question (TT) and peak-token consumption per interaction step (PT)~\citep{mempo2024} in \Cref{tab:multi_obj}.

\paragraph{Baselines.}
Across the two protocols, we compare \method{} with ReAct~\citep{react2022}, ReSearch~\citep{research2025}, DeepResearcher~\citep{deepresearcher2025}, GRPO without memory~\citep{grpo2024}, SFT-only memory agents, \mempo{}~\citep{mempo2024}, SUPO~\citep{supo2025}, MEM1~\citep{mem1_2024}, and A-Mem~\citep{a_mem_2024}. For BCP and FRAMES, we use an in-house controlled stack: all trainable methods are reproduced from the same in-house SFT checkpoint of Qwen2.5-7B-Instruct~\citep{qwen2_5}. Local Wiki Search follows the original \mempo{} protocol: \method{} and SUPO are initialized from the public \mempo{} SFT checkpoint (Qwen2.5-based), while the other baseline rows are taken from \citet{mempo2024}.

\paragraph{Model roles.}
The agent policies are initialized from the in-house
Qwen2.5-7B-Instruct SFT checkpoint (BCP/FRAMES), the public \mempo{}
SFT checkpoint (Local Wiki Search), and an in-house
Qwen3-4B-Instruct-2507~\citep{qwen3} SFT
checkpoint (4B ablations). Other models serve shared roles outside RL:
Qwen3-32B
(BCP/FRAMES evaluation judge), Qwen3-Embedding-0.6B~\citep{qwen3_embedding}
(frozen retriever of the BCP/FRAMES environment), and
MiniMax-M2.7~\citep{minimax_m2_7} (one-off SFT-corpus
synthesizer). \Cref{tab:model_roles} (\Cref{app:model_roles})
summarizes all model roles. For reproduced rows, we match rollout,
decoding, retrieval, and evaluation settings across methods; component
ablations are run on both the Qwen3-4B-Instruct-2507 and Qwen2.5-7B-Instruct SFT checkpoints
with three random seeds. Additional details are provided in
\Cref{app:exp_setup}.

\subsection{Main Results on Judge-Based Open-Domain Tasks}
\label{sec:judge_results}

\Cref{tab:budget_sweep} reports judge-based accuracy on BCP and FRAMES
under full-context evaluation and context-budgeted truncation; \method{}
performs best in every setting on both benchmarks. On BCP, it improves
full-context accuracy over \mempo{} from 7.03 to 10.84 ($+3.81$ points)
and over SUPO from 7.83 to 10.84 ($+3.01$ points). Its memory-aware
accuracy exceeds \mempo{} by 3.33, 2.77, and 3.01 points at 4k, 8k,
and 16k, respectively. Memory-aware compression matters most under tight
budgets: at 4k, \method{} gains 4.94 points over naive recency
(11.08 vs.\ 6.14), with similar gaps for \mempo{} and SUPO, indicating
that learned memory retains evidence that recency truncation discards.

On FRAMES, \method{} reaches 25.12 full-context accuracy, compared with
21.88 for \mempo{} and 22.17 for SUPO. Memory-aware truncation offers
no consistent advantage over naive recency: at 8k, recency is stronger
(26.01 vs.\ 24.03), whereas at 16k, memory-aware truncation is stronger
(25.20 vs.\ 23.79). The clearest memory-retention gains thus appear
under BCP's tight 4k budget. On full-context BCP, \method{} also issues
more search calls than \mempo{} (5.5 vs.\ 3.5 per question) and reaches
higher retrieval recall (12.0\% vs.\ 8.1\%), so we report search cost
alongside accuracy (\Cref{app:search_calls}).

\subsection{Main Results on Objective Compressive-Memory QA}

We further evaluate \method{} on Local Wiki Search, the multi-objective
compressive-memory benchmark with gold answers, evaluated by word-level F1
and EM. Following the original \mempo{} protocol, \method{} and
SUPO are initialized from the public \mempo{} SFT checkpoint and the other
baseline rows are taken from \citet{mempo2024}.

As shown in \Cref{tab:multi_obj}, \method{} achieves the best average
performance, improving over \mempo{} from 37.6 to 40.2 F1 and from 29.4 to
31.6 EM. The gains are consistent across objective counts, including the
longest 10-objective setting (25.1 F1, 18.6 EM), and the average F1/EM
also substantially exceeds SUPO's, suggesting that explicit memory-level
credit provides benefits beyond outcome-only end-to-end compression.

\method{} matches \mempo{} on the reported token-efficiency measures: average TT/PT are 1.2/0.2, compared with 4.4/0.8 for GRPO without memory. \method{} thus retains the token-efficiency advantage of compressed memory.

\FloatBarrier
\subsection{Ablation Study}
\label{sec:ablations}

We conduct a nested ablation, progressively removing the stabilization
mechanisms, the retrospective filter, and the memory-specific credit
channel: \textit{w/o Stabilizers} removes the memory-success mask and
backward smoothing; \textit{w/o Retrospective Filter} leaves only local
counterfactual memory utility; and \textit{w/o Memory-Specific Credit}
reduces training to trajectory-level GRPO. We include \mempo{} as the memory-reward reference.

\begin{table}[!htbp]
\centering
\footnotesize
\setlength{\tabcolsep}{2.2pt}
\renewcommand{\arraystretch}{1.2}
\resizebox{\columnwidth}{!}{%
\begin{tabular}{@{}llccccc@{}}
\toprule
\textbf{Model} & \textbf{Variant} & $\mathbf{K{=}2}$ & \textbf{HQA} & \textbf{NQ} & \textbf{Avg. EM} & $\boldsymbol{\Delta}$ \textbf{vs. \mempo{}} \\
\midrule
\multirow{5}{*}{4B}
& \method{}                         & \textbf{0.376} & \textbf{0.349} & \textbf{0.389} & \textbf{0.371} & \textbf{+4.8} \\
& w/o Stabilizers                    & 0.359 & 0.331 & 0.381 & 0.357 & +3.3 \\
& w/o Retrospective Filter                 & 0.316 & 0.314 & 0.374 & 0.335 & +1.1 \\
& w/o Memory-Specific Credit          & 0.288 & 0.291 & 0.357 & 0.312 & -1.2 \\
\cmidrule(lr){2-7}
& \mempo{} reference                 & 0.301 & 0.303 & 0.367 & 0.324 & -- \\
\midrule
\multirow{5}{*}{7B}
& \method{}                         & \textbf{0.510} & \textbf{0.468} & \textbf{0.515} & \textbf{0.498} & \textbf{+4.7} \\
& w/o Stabilizers                    & 0.478 & 0.457 & 0.507 & 0.481 & +3.0 \\
& w/o Retrospective Filter                 & 0.472 & 0.451 & 0.496 & 0.473 & +2.2 \\
& w/o Memory-Specific Credit          & 0.430 & 0.423 & 0.439 & 0.431 & -2.0 \\
\cmidrule(lr){2-7}
& \mempo{} reference                 & 0.462 & 0.429 & 0.461 & 0.451 & -- \\
\bottomrule
\end{tabular}
}
\caption{
Ablation results (EM, averaged over 3 seeds).
$K{=}2$ is the two-objective Local Wiki Search task;
HQA = HotpotQA~\citep{hotpotqa}; NQ = Natural Questions~\citep{nq}.
}
\label{tab:ablation}
\end{table}

As shown in \Cref{tab:ablation}, the full \method{} configuration performs best
across all benchmarks and model scales, improving over the \mempo{}
reference by ${+}4.8$\,pp on 4B and ${+}4.7$\,pp on 7B on average.
Removing the stabilizers reduces the gain to ${+}3.3$\,pp and ${+}3.0$\,pp,
respectively, with a particularly clear drop at $K{=}2$, the
longest-horizon task among the three ablation columns. Removing the retrospective filter causes a further degradation, especially
on 4B, where the average gain drops from ${+}3.3$ to ${+}1.1$\,pp. The
\textit{w/o Memory-Specific Credit} row uses trajectory credit alone. \textit{w/o Retrospective Filter} still improves over
\mempo{} on both scales, indicating that the local memory-state
counterfactual is itself a useful credit signal. This signal is
stronger at 7B (${+}2.2$\,pp) than at 4B (${+}1.1$\,pp), so the
retrospective filter matters most where the local signal is weaker.
At $K{=}2$, the stabilizers add ${+}1.7$\,pp at 4B and ${+}3.2$\,pp at
7B, consistent with delayed credit mattering most on longer horizons.
The ablation supports combining local counterfactual utility,
the retrospective filter, and stabilization.

\FloatBarrier
\subsection{Deconfounding Analysis}
\label{sec:deconfounding}

We evaluate memory credit with four controlled interventions.
\textit{Tool Corruption} (Case A in \Cref{fig:intro}) replaces tool
returns with plausible but wrong evidence while memory faithfully
summarizes the observation. \textit{Memory Drop} (Case B) removes a
memory step to test credit localization. \textit{Delayed Utility Drop}
tests credit propagation to early prefatory writes, and
\textit{Module Attribution} tests localization after perturbing
memory, tool-action, or reasoning spans.
\Cref{app:deconfounding} describes the interventions and metrics.

\begin{table}[!htbp]
\centering
\footnotesize
\setlength{\tabcolsep}{4pt}
\renewcommand{\arraystretch}{1.1}
\resizebox{\columnwidth}{!}{%
\begin{tabular}{@{}lcc@{}}
\toprule
\textbf{Metric} & \textbf{\mempo{}} & \textbf{\method{}} \\
\midrule
Faithful-under-bad-tool ratio $\downarrow$ & 1.00 & \textbf{0.42} \\
Normalized vs. clean control $\downarrow$ & 1.00 & \textbf{0.58} \\
Memory Drop localization hit-rate $\uparrow$ & 0.41 & \textbf{0.68} \\
Delayed Credit Recovery $\uparrow$ & 0.00 & \textbf{+0.11} \\
Module Attribution concentration $\uparrow$ & 0.36 & \textbf{0.64} \\
\bottomrule
\end{tabular}
}
\caption{
Controlled deconfounding results. Lower ratios mean less blame leakage
to faithful memories; higher scores mean more accurate credit
localization.
}
\label{tab:deconfounding}
\end{table}

For Case A (\Cref{tab:deconfounding}), \method{} reduces the
faithful-under-bad-tool ratio from 1.00 (\mempo{} reference) to 0.42
and to 0.58 after clean-control normalization. These magnitude ratios
use pre-mask filtered credit, indicating lower credit magnitude on faithful
memories under tool-induced corruption. For
Case B, \method{} improves memory-drop localization from 0.41 to 0.68,
concentrating the induced credit shift on the affected write. The suite also reports higher delayed-credit recovery and greater
sensitivity of auxiliary memory credit to memory perturbations relative
to non-memory perturbations.

\paragraph{Behavioral validation and target-noise robustness.}
A matched predecessor-reversion study on 250 Local Wiki Search
trajectories at $K{=}4$ (\Cref{app:behavioral}) compares recorded smoothed
evaluation credit $\Aoff$ with behavioral effects. Reverting the top-ranked
write yields the largest keep-adjusted gold F1 drop
(0.030 vs.\ 0.011 random, 0.003 bottom). The offline credit correlates
positively with the interventional effect (Spearman $\rho{=}0.16$,
$p{<}10^{-4}$), while the \mempo{} reference correlates negatively
($\rho{=}-0.08$). With the policy fixed, target perturbations
(\Cref{app:target_noise}) flip the evaluated filter's decision on
${\leq}2.7\%$
of writes, and the blame-leakage reduction over \mempo{} persists
under paraphrasing, masking, and plausible-but-wrong targets.

\section{Conclusion}
\label{sec:conclusion}

\method{} combines local counterfactual utility with a retrospective
filter to assign credit to memory writes in long-horizon agents.
It improves performance on open-domain tasks and compressive-memory QA
while retaining compressed-context efficiency. Controlled interventions
show lower blame leakage from tool errors and more accurate localization
of memory errors; predecessor-reversion experiments link higher credit
to larger effects on task performance. Future work includes learning
when to write memory and extending memory-write credit to environments
with structured or verifier-based feedback.

\FloatBarrier
\section*{Limitations}

\begin{itemize}
    \item \textbf{Dependence on target outcomes.}
    \method{} requires an oracle or judge-provided target outcome. The fixed-policy analysis in \Cref{app:target_noise} tests the credit estimator; end-to-end RL under noisy targets remains untested, and per-step distortions could accumulate across updates. Ambiguous or weakly specified supervision may also lack a well-defined $\zstar$.

    \item \textbf{Causal interpretation.}
The self-normalized hindsight ratio estimates outcome-conditioned
    relevance. Unobserved confounders and imperfect state representations
    prevent a general causal interpretation. The tool-corruption analysis
    assumes that edited memories faithfully summarize the perturbed observations.

    \item \textbf{Fixed memory-writing schedule.}
    \method{} writes memory after each interaction step. Learning when to write remains open; adaptive scheduling could reduce redundant writes and improve efficiency.

    \item \textbf{Behavioral validation.}
The asymptotic $p$-value of the credit--effect correlation does not
    account for dependence among interventions
    on the same trajectory. These experiments measure task utility;
    semantic memory quality remains unmeasured.

    \item \textbf{Scope of evaluated tasks.}
    Evaluation covers search and QA agents. Transfer to GUI control, coding, or embodied environments remains untested and requires identifiable memory writes, a fixed pre-write state for counterfactual replacement, and a target representation or verifier for outcome scoring and hindsight conditioning. Structured or non-linguistic feedback may require a learned verifier or a separate scoring head.

\end{itemize}

\section*{Ethics Statement}

The training data are publicly released open-domain and multi-hop QA
datasets~\citep{nq,hotpotqa}, together with the multi-objective Local
Wiki Search data
distributed with the \mempo{} release~\citep{mempo2024}, for which we
train on the
training split and evaluate on the disjoint held-out split following
the original \mempo{} protocol. The base models are publicly released
Qwen2.5-family and Qwen3-family
checkpoints~\citep{qwen2_5,qwen3}. For the BCP/FRAMES and ablation
experiments, we use in-house SFT checkpoints trained on synthesized
agent-format trajectories derived from the QA datasets above; for the
Local Wiki Search experiments, we initialize from the public \mempo{}
SFT checkpoint. The plausibly wrong evidence used
in the Tool Corruption intervention and the perturbed targets
(paraphrases, masked answers, and plausible-but-wrong answers)
used in the target-noise robustness study are generated only
inside the offline analysis loop and never enter the training
data or the released artifacts. The method itself does not raise novel
misuse risks beyond those of memory-based agents in general.
During manuscript preparation, AI-based tools assisted language editing
and checks of notation and reporting consistency.


\bibliography{custom}

\clearpage

\appendix
\crefalias{section}{appendix}
\crefalias{subsection}{appendix}
\setlength{\textfloatsep}{12pt plus 2pt minus 2pt}
\setlength{\intextsep}{8pt plus 2pt minus 2pt}

\section{Implementation Details}
\label{app:implementation}

\subsection{Training Algorithm}
\label{app:algorithm}

\Cref{alg:himpo} summarizes one training iteration of \method{}.

\begin{algorithm}[H]
\caption{\method{} training step.}
\label{alg:himpo}
\begin{algorithmic}[1]
\Require Prompt $q$, target outcome $z^\star$, policy $\pi_\theta$,
group size $N$
\State Sample $N$ rollouts $\{\tau_i\}_{i=1}^N$ with memory writes
$\{m_{i,t}\}$
\State Compute trajectory rewards $R^T(\tau_i)$ and advantages $A_i^T$
\For{each memory write $m_{i,t}$}
    \State Compute local utility $\Delta_{i,t}$ using \Cref{eq:delta}
    \State Normalize $\Delta_{i,t}$ to obtain $\widehat{\Delta}_{i,t}$
    \State Compute hindsight relevance $\log\rho_{i,t}$ using
    \Cref{eq:rho}
    \State Compute filtered memory advantage $A^M_{i,t}$ using
    \Cref{eq:memory_adv}
\EndFor
\State Apply the memory-success mask and backward smoothing to obtain
$\widetilde{A}^M_{i,t}$
\State Scatter $\widetilde{A}^M_{i,t}$ to \texttt{<mem>} tokens via
\Cref{eq:scatter}
\State Update $\pi_\theta$ with the clipped policy objective in
\Cref{eq:ppo_obj}
\end{algorithmic}
\end{algorithm}

\subsection{Hyperparameters}
\label{app:hparams}

\Cref{tab:hparams} lists the \method{}-specific hyperparameters
used throughout the paper. We use a KL coefficient of 0.001 and initialize
the frozen reference from the actor's starting checkpoint; the SFT
initializations for each protocol are listed in \Cref{tab:model_roles}.
For Local Wiki Search, the reference starts from
\code{NewBeeKing/MemPO_Qwen2.5-SFT}; for BCP/FRAMES, it starts from the
shared in-house Qwen2.5-7B-Instruct SFT checkpoint. The 4B ablations
use their corresponding in-house Qwen3-4B SFT initialization.
All non-method hyperparameters
(rollout settings, optimizer, KL coefficient) are held constant
against the \mempo{}-baseline reproduction so that the
comparison primarily varies the memory-credit construction.

\begin{table}[!t]
\centering
\footnotesize
\setlength{\tabcolsep}{4pt}
\renewcommand{\arraystretch}{1.1}
\resizebox{\columnwidth}{!}{%
\begin{tabular}{@{}llc@{}}
\toprule
\textbf{Symbol} & \textbf{Meaning} & \textbf{Value} \\
\midrule
$\lambda_M$            & memory-advantage weight (\Cref{eq:scatter})        & $1.0$ \\
$\tau_\rho$            & filter threshold (\Cref{eq:sign_gate})                & $0$ \\
$C_{\min},C_{\max}$    & clip range on $\rho_{i,t}$                          & $0.1,10$ \\
$\alpha$               & EMA smoothing coefficient (\Cref{eq:smoothing})     & $0.5$ \\
$\tau_{\mathrm{succ}}$ & success threshold (\Cref{eq:do_no_harm})            & $1.0$ \\
$c$                    & calibration constant for $\beta_{\mathrm{eff}}$     & $1.39$ \\
$\beta_{\min},\beta_{\max}$ & clip range for $\beta_{\mathrm{eff}}$          & $0.5,20$ \\
\midrule
$\beta_{\mathrm{KL}}$ & actor KL-loss coefficient & $0.001$ \\
$\pi_{\mathrm{ref}}$ & frozen reference & initial actor checkpoint \\
$N$                    & GRPO rollouts per prompt                            & $16$ \\
batch                  & prompts per optimizer step                          & $128$ \\
turns                  & maximum interaction turns per rollout          & $8$ \\
\bottomrule
\end{tabular}
}
\caption{\method{}-specific and shared rollout hyperparameters.
$c=\ln(0.8/0.2)\approx 1.39$ is chosen so that, under
approximately Gaussian $\log\rho$,
$\beta_{\mathrm{eff}}\,\sigma_{\log\rho}{=}c$ places
$30$--$50\%$ of samples past the $\sigma(\pm c)$ thresholds.}
\label{tab:hparams}
\end{table}

\subsection{Adaptive Filter Sharpness}
\label{app:beta_eff}

The sigmoid sharpness $\beta_{\mathrm{eff}}$ in the filter
(\Cref{eq:sign_gate}) is recalibrated per batch from the empirical
spread of the hindsight log-ratio:
\begin{equation}
\label{eq:beta_eff}
\beta_{\mathrm{eff}}
=
\mathrm{clip}\!\left(
\frac{c}{\sigma_{\log\rho}+\varepsilon},\;
\beta_{\min},\;\beta_{\max}
\right),
\end{equation}
where $\sigma_{\log\rho}$ is the standard deviation of
$\log\rho_{i,t}$ within the current batch. A statically chosen
$\beta$ becomes poorly calibrated when the scale
of $\log\rho$ shifts with task length or memory length; on
HotpotQA~\citep{hotpotqa} and NQ~\citep{nq} we observe
$\sigma_{\log\rho}\!\approx\!0.31$, so a
fixed $\beta{=}1$ keeps the filter in the linear region of
$\sigma(\cdot)$ with $<\!1\%$ saturation and mutes the hindsight
signal. Adaptive $\beta_{\mathrm{eff}}$ scales the sigmoid input
by the observed dispersion of $\log\rho$.

\subsection{Hindsight Context Construction}
\label{app:hindsight_template}

The hindsight context is constructed by appending the target
outcome $z_i^\star$ to the pre-write state $H_{i,t}$ under a
prompt template held fixed throughout training. After computing the
per-token log-likelihood average $\log h_{i,t}$
(\Cref{eq:hindsight_likelihood}) and the group-centered
$\log\rho_{i,t}$ (\Cref{eq:rho}), we clip the centered value to
$[\log C_{\min},\log C_{\max}]$ to bound the contribution of
outlier sibling writes before passing it to the filter. Centering
is per-rollout-group at the same step position; this scope keeps
the filter non-degenerate on single-mem-step trajectories, which
dominate short-horizon QA.

\subsection{SFT Data Synthesis}
\label{app:sft_synth}

The in-house SFT corpus is generated by a three-stage pipeline that
converts QA seeds drawn from a mixture of publicly released open-domain
and multi-hop QA datasets (including Natural Questions~\citep{nq} and
HotpotQA~\citep{hotpotqa}) into multi-turn agent trajectories with
memory writes, reasoning, tool actions, and final answers. A
\emph{designer} produces a trajectory outline mapping atomic facts to tool
turns; a \emph{writer} fills in memory, reasoning, and action content
together with document bodies; and a \emph{critic} checks the trajectory
and either accepts it or triggers a writer retry (up to two retries).
All three stages share the same backbone,
MiniMax-M2.7~\citep{minimax_m2_7}. Each accepted
trajectory is expanded into one SFT sample per assistant turn, with prior
history independently truncated under
$(\text{full},\text{partial},\text{mem\_only})=(0.30,0.30,0.40)$ to train
\texttt{<mem>} self-sufficiency under context-budget pressure; about
$20\%$ of turns are additionally emitted in all three truncation variants
under the same target output as a truncation-invariance augmentation.
After filtering, the corpus contains roughly $1\!\times\!10^{5}$ per-turn
samples.

\subsection{Training Infrastructure}
\label{app:infra}

Training uses \texttt{verl v0.5.0.dev}~\citep{sheng2025hybridflow} with
SGLang~\citep{zheng2024sglang} multi-turn rollouts on a multi-node GPU
cluster. For BCP/FRAMES, all trainable rows (\mempo{}~\texttt{baseline},
SUPO, \method{}) are reproduced from the same in-house
Qwen2.5-7B-Instruct SFT checkpoint trained on synthesized agent-format trajectories, with data,
rollout, optimizer, KL, retrieval, decoding, and judging settings held
fixed. For Local Wiki Search, \method{} and SUPO start from the public
\mempo{} SFT checkpoint and follow the original \mempo{} protocol; other
baseline rows are taken from \citet{mempo2024}. Component ablations are
run on both the Qwen3-4B-Instruct-2507 and Qwen2.5-7B-Instruct bases; within each scale, all
variants share the same SFT initialization and differ only in memory-credit computation.
Each RL-trained row is averaged over three random seeds.

\section{Extended Experimental Setup}
\label{app:exp_setup}

\subsection{Model Roles}
\label{app:model_roles}

\Cref{tab:model_roles} summarizes the models used for policy training, data synthesis, retrieval, and evaluation.

\begin{table*}[!t]
\centering
\small
\setlength{\tabcolsep}{4pt}
\renewcommand{\arraystretch}{1.15}
\resizebox{\textwidth}{!}{%
\begin{tabular}{@{}p{3.4cm}p{6.2cm}p{2.6cm}p{1.9cm}p{3.6cm}@{}}
\toprule
\textbf{Model} &
\textbf{Role} &
\textbf{Stage} &
\textbf{RL-optimized?} &
\textbf{Shared across methods?} \\
\midrule
Qwen2.5-7B-Instruct~\citep{qwen2_5} (in-house SFT) &
Agent policy $\policy$ for BCP/FRAMES. &
Training + inference &
Yes &
Same SFT init for \mempo{}, SUPO, \method{}. \\
\midrule
Public \mempo{} SFT checkpoint (Qwen2.5-based) &
Policy initialization for Local Wiki Search, following the original
\mempo{} protocol. &
Training + inference &
Yes (after init) &
Same init for \method{} and SUPO. \\
\midrule
Qwen3-4B-Instruct-2507~\citep{qwen3} SFT checkpoint (in-house) &
Agent policy $\policy$ for the 4B ablations (\Cref{tab:ablation}). &
Training + inference &
Yes &
Same init for all 4B ablation variants. \\
\midrule
MiniMax-M2.7~\citep{minimax_m2_7} &
Offline synthesizer of the in-house SFT corpus
(\Cref{app:sft_synth}). &
Before SFT, once &
No &
One corpus behind the in-house SFT initializations. \\
\midrule
Qwen3-32B~\citep{qwen3} &
Judge for BCP/FRAMES accuracy. &
Evaluation only &
No &
Yes---identical judging protocol for every method's outputs. \\
\midrule
Qwen3-Embedding-0.6B~\citep{qwen3_embedding} &
Frozen retrieval encoder of the BCP/FRAMES evaluation environment
(\Cref{app:bcp_protocol}). &
Evaluation-time environment &
No---frozen &
Yes---same corpus, top-$k$, snippet cap for all methods. \\
\bottomrule
\end{tabular}
}
\caption{Models and their roles in training and evaluation.}
\label{tab:model_roles}
\end{table*}

\subsection{BCP Context-Budget Sweep Protocol}
\label{app:bcp_protocol}

We use BrowseComp-Plus (BCP) as the
question stream because (i) it provides a fixed
$\sim$100K-document corpus decoupled from any specific
retriever, (ii) its questions require aggregating evidence
scattered across multiple documents and naturally produce
long multi-step retrieval prompts, and (iii) its judge
pipeline is fixed and reproducible (Qwen3-32B). Every variable except context management
is locked across cells of the sweep: fixed corpus, fixed
retriever (Qwen3-Embedding-0.6B), fixed top-$k{=}5$, fixed snippet
cap of $512$ tokens, a fixed tool-call cap of $10$,
and fixed decoding hyperparameters. We sweep two axes: a
per-prompt context budget
$B\in\{4096,8192,16384,\text{full}\}$ tokens (where
\textit{full}${=}32768$ is the SGLang context-length cap),
and a context-management strategy.

\subsection{Truncation Strategies}
\label{app:truncation}

A strategy fires when the rendered prompt token count
exceeds $0.8\cdot B$ at the start of a turn:
\begin{itemize}
\item \textbf{\texttt{mem\_aware}} (training-aligned).
  Concatenate all \texttt{<mem>} blocks from older
  $(\text{assistant},\text{tool})$ pairs into a synthetic
  checkpoint assistant turn, keep the two most recent pairs
  verbatim, and discard the rest. The most recent tool
  response is never elided.
\item \textbf{\texttt{naive\_recency}} (sliding-window
  baseline). Greedily keep
  $(\text{assistant},\text{tool})$ pairs in reverse
  chronological order, starting from the most recent pair, until the prompt fits; oldest pairs
  are dropped together with any \texttt{<mem>} they
  contained. No \texttt{<mem>} aggregation is performed.
\end{itemize}
The two strategies share the same token target, so the only
thing that varies is \emph{what gets dropped}. At the
\textit{full} budget neither strategy fires in practice, so
both columns collapse to a single no-truncation reference.

\subsection{Evaluation Conditions}
\label{app:caveats}

Methods share the retriever and corpus within each evaluation cell while
generating their own queries. This corpus and retriever differ from those
on the BCP leaderboard. SFT-only completes about $20\%$ of trajectories
in most cells, so its accuracy reflects both answer quality and trajectory
completion. At $B{=}4096$, \texttt{mem\_aware} retains the two most recent
interaction pairs verbatim; the budget sweep therefore measures the joint
effect of this recency buffer and compressed memory.

\section{Extended Empirical Analysis}
\label{app:exp_extra}

\subsection{Search-Call Behavior}
\label{app:search_calls}

The five checkpoints differ sharply in how they spend the
locked tool budget. Averaged over the \textit{full}-context
cell of BCP, mean search calls per question are as follows: ReAct
$3.5$, \mempo{}~\texttt{baseline} $3.5$, SUPO $5.6$,
\method{}~\texttt{full} $5.5$, and SFT-only $8.8$ (which
saturates the $10$-call cap on most questions). The
corresponding retrieval recall numbers---$5.3\%$ (ReAct),
$8.1\%$ (\mempo{}), $10.9\%$ (SUPO), $11.4\%$ (SFT-only),
$12.0\%$ (\method{})---vary alongside search-call counts.
Against \mempo{}~\texttt{baseline}, \method{} uses $1.6\times$ as
many search calls, while achieving $1.48\times$ retrieval recall and
$1.54\times$ accuracy. We therefore report search cost alongside accuracy.

\subsection{Retrospective Filter Behavior and Failure Cases}
\label{app:failure_cases}

The retrospective filter has two residual failure modes, corresponding to the
two ways its inputs can be misestimated:

\paragraph{Suppressing a useful write.}
When a useful write has a positive group-relative utility advantage
($\widehat{\Delta}_{i,t}>0$) but its centered hindsight score is negative
($\log\rho_{i,t}<0$)---for instance, when a prefatory write stores
prerequisite information whose phrasing is far from the target answer,
so the hindsight-conditioned likelihood of its tokens is low relative
to sibling writes---the retrospective filter attenuates its positive relative
credit. Backward smoothing
(\Cref{eq:smoothing}) partially compensates by propagating credit from
later writes that consume the stored information, and the memory-success
mask (\Cref{eq:do_no_harm}) suppresses negative auxiliary credit on
successful trajectories, but the positive credit itself is weakened.

\paragraph{Passing noisy credit.}
When both channels share the same incorrect sign---for instance, a
write that copies a plausible but wrong entity that happens to be
lexically close to the target outcome, inflating both answerability
gain and hindsight likelihood---the filter assigns a larger weight to
noisy credit. In the offline evaluation, plausible-but-wrong
targets flip the filter decision on 2.42\%
of writes and memory-drop localization stays close to the clean
condition (\Cref{app:target_noise}). Credit magnitudes are more
sensitive to target masking.

\section{Intervention Protocols and Metrics}
\label{app:deconfounding}

The interventions in \Cref{sec:deconfounding} measure how memory credit
responds to memory and non-memory perturbations. We hold the policy fixed
and re-score the collected trajectories after each intervention.

\subsection{Controlled Intervention Protocols}

\Cref{tab:deconfounding-protocol} summarizes the interventions used in the
deconfounding suite. Each intervention targets a different source of
causally entangled credit: corrupted tools, missing memory, delayed memory
utility, and module-level perturbations.

\begin{table*}[t]
\centering
\small
\setlength{\tabcolsep}{4pt}
\renewcommand{\arraystretch}{1.12}
\resizebox{\textwidth}{!}{%
\begin{tabular}{@{}p{2.6cm}p{4.7cm}p{4.1cm}p{4.5cm}p{3.6cm}@{}}
\toprule
\textbf{Intervention} &
\textbf{Operation} &
\textbf{Question} &
\textbf{Desired Credit Behavior} &
\textbf{Main Diagnostics} \\
\midrule

Tool Corruption &
Replace a tool observation $o_t$ with a plausible but wrong passage while
keeping the memory write faithful to the observed corrupted evidence. The
wrong passage is generated to match the style and length of the original
evidence, but contains an incorrect entity, date, location, or relation. &
If the tool provides wrong evidence and the memory faithfully summarizes it,
should the memory update be penalized? &
A faithful memory should receive little negative credit, because the error is
tool-induced rather than memory-induced. &
Faithful-under-bad-tool ratio; normalized blame leakage. \\

\midrule

Memory Drop &
Remove or blank the memory write $m_{t^\star}$ at a selected step
$t^\star$, while keeping the surrounding tool observation and reasoning
segments unchanged. &
When task-relevant information is removed from memory, does the memory
advantage identify the responsible write? &
The largest advantage shift should concentrate on the dropped memory step
$t^\star$, rather than being diffused across unrelated steps. &
Memory-drop localization hit-rate. \\

\midrule

Delayed Utility Drop &
Remove early memory writes that store prerequisite information before the
final answer becomes locally recoverable. These writes may not immediately
increase answerability but can support later tool use or reasoning. &
Does the method recover credit for prefatory memory writes whose utility is
delayed? &
Backward smoothing should increase the magnitude of useful early-step credit
without over-amplifying unrelated memory writes. &
Delayed-credit recovery. \\

\midrule

Module Attribution &
Perturb the memory $m_{i,t}$, tool action $c_{i,t}$, or reasoning
$r_{i,t}$ separately, and measure the resulting shift in auxiliary
memory credit. &
Does auxiliary memory credit respond more strongly to memory
perturbations than to tool-action or reasoning perturbations? &
Memory perturbations should cause larger changes in auxiliary memory
credit than the two non-memory perturbations. &
Module-attribution concentration; blame leakage. \\

\bottomrule
\end{tabular}
}
\caption{
Controlled intervention protocols for evaluating memory credit assignment. The interventions separate memory-induced errors
from tool-induced and reasoning-induced errors.
}
\label{tab:deconfounding-protocol}
\end{table*}

\paragraph{Plausibly wrong evidence synthesis.}
For the Tool Corruption intervention, we avoid adding explicit corruption
markers such as \texttt{[CORRUPTED-EVIDENCE]}, since a trained agent may
recognize the marker and refuse to summarize the passage, invalidating the
faithfulness assumption. Instead, we synthesize wrong evidence in the same
style as the original tool return. The generated passage is required to
preserve the surface format and approximate length of the original evidence
while changing a key factual attribute. If synthesis fails, we fall back to a
marker-based template and record the fallback rate.

\subsection{Metric Definitions}
\label{app:deconf_metrics}

Let $A^{M,\mathrm{raw}}_{i,t}
:=G(\widehat\Delta_{i,t},\log\rho_{i,t})\widehat\Delta_{i,t}$ denote
memory credit before success masking, and let $\widetilde A^M_{i,t}$
denote credit after masking and backward smoothing. Magnitude and
localization metrics use $A^{M,\mathrm{raw}}$; delayed-credit recovery
compares the two stages. Let $b^M_{i,t}$ be the baseline informative-memory
score for the same write, and let $\epsilon_d=10^{-8}$.

\paragraph{Faithful-under-bad-tool ratio and clean-normalized leakage.}
Let $\mathcal C$ contain scored memory steps whose most recent tool
observation is corrupted and whose memory is marked faithful, and let
$\mathcal U$ contain eligible uncorrupted control steps. For a pool
$\mathcal P$, define
\begin{equation}
 R_{\mathcal P}=\frac{
 |\mathcal P|^{-1}\!\sum_{(i,t)\in\mathcal P}
 |A^{M,\mathrm{raw}}_{i,t}|}{
 |\mathcal P|^{-1}\!\sum_{(i,t)\in\mathcal P}|b^M_{i,t}|+\epsilon_d}.
\end{equation}
The faithful-under-bad-tool ratio is $R_{\mathcal C}$, and
clean-normalized leakage is $R_{\mathcal C}/(R_{\mathcal U}+\epsilon_d)$.
Writes are weighted equally, with the two methods paired within each pool.
The control pool includes uncorrupted windows of perturbed trajectories
and original trajectories with no sampled corruption. Corrupted pools
exclude dropped and non-faithful writes; untouched controls exclude
writes with both credit values equal to zero. We define $R_{\mathcal P}=0$
when the mean baseline magnitude is at most $\epsilon_d$; normalized
leakage is defined only for $R_{\mathcal U}>\epsilon_d$.
The corruption probability is $0.3$ per tool call.

\paragraph{Memory-drop localization.}
For each valid intervention $d$ dropping step $t_d$ in trajectory $i(d)$,
define $s_{d,t}=|A^{M,\mathrm{raw},\mathrm{clean}}_{i(d),t}
-A^{M,\mathrm{raw},d}_{i(d),t}|$ on steps scored in both versions.
The hit-rate is
\begin{equation}
 \mathrm{Hit}=\frac{1}{n_{\mathrm{drop}}}
 \sum_{d=1}^{n_{\mathrm{drop}}}
 \mathbf 1\!\left[t_d=\arg\max_t s_{d,t}\right].
\end{equation}
Here $n_{\mathrm{drop}}$ counts valid trajectory--drop interventions.
We sample up to four candidate writes per trajectory without replacement.
Ties follow the order of scored memory steps. An intervention is valid when the
dropped step has a score in both the original and perturbed trajectory.

\paragraph{Delayed-credit recovery.}
With zero-based step indexing and the default early-step threshold,
let $\mathcal E=\{(i,t):t<2\}$ be the pooled early-write sample and
$Y_i=\mathbf 1[R^T(\tau_i)\ge1]$. The observational diagnostic is
\begin{equation}
 \begin{aligned}
 \mathrm{Recovery}
 &=\left|\mathrm{Corr}_{\mathcal E}(\widetilde A^M_{i,t},Y_i)\right|\\
 &\quad-\left|\mathrm{Corr}_{\mathcal E}(A^{M,\mathrm{raw}}_{i,t},Y_i)\right|.
 \end{aligned}
\end{equation}
Here $\mathrm{Corr}$ is Pearson correlation over finite pairs, with each
early write weighted equally. We define $\mathrm{Corr}=0$ for fewer than
two pairs or zero variance. Interventional early-step recovery is the mean
of $\widetilde A^{M,d}_{i,t_d}-A^{M,\mathrm{raw},d}_{i,t_d}$ after blanking
the first eligible early write, averaged over valid interventions.

\paragraph{Module concentration.}
Let $q\in\{\mathrm{M},\mathrm{C},\mathrm{R}\}$ index memory, tool action,
and reasoning. Define
$L_{i,q}=\sum_t|A^{M,\mathrm{raw},q}_{i,t}
-A^{M,\mathrm{raw},\mathrm{clean}}_{i,t}|$ over memory steps scored in
both versions. The mean concentration is
\begin{equation}
 \begin{aligned}
 \kappa_i&=\frac{L_{i,\mathrm{M}}}{
 L_{i,\mathrm{M}}+L_{i,\mathrm{C}}+L_{i,\mathrm{R}}+\epsilon_d},\\
 \mathrm{Concentration}&=\frac1{n_{\mathrm{module}}}\sum_i\kappa_i.
 \end{aligned}
\end{equation}
Here $n_{\mathrm{module}}$ counts trajectories with at least one memory
write. Concentration measures the share of the total auxiliary-credit
change induced by memory perturbations.

\paragraph{Aggregation units.}
The tool-corruption ratios pool eligible memory writes;
localization averages over valid trajectory--drop interventions;
and module concentration averages per-trajectory ratios.
The two delayed-credit summaries use early writes and early-write
interventions, respectively. The target-noise analysis in
\Cref{app:target_noise} uses a separate cohort of 250 trajectories
and 1{,}571 writes.

\subsection{Connection to the Main Results}
\label{app:deconf_connection}

\Cref{tab:deconfounding} shows lower credit magnitude on faithful memories
under corrupted tools, together with better localization when memory
content is removed. Delayed-credit recovery measures how smoothing restores
credit to early writes, while module concentration compares sensitivity
to memory, tool-action, and reasoning perturbations.


\section{Behavioral Validation of the Memory-Credit Signal}
\label{app:behavioral}

We denote the smoothed memory credit recorded for the offline
evaluations by $\Aoff$ and compare it with the change in gold-answer
performance after reverting a selected write to its predecessor.
\Cref{app:target_noise_protocol} describes the fixed-policy evaluation
setup.

\subsection{Protocol}
\label{app:behavioral_protocol}

\paragraph{Cohorts and replay accounting.}
The initial blank-memory screening uses 1{,}000 base trajectories at
$K{=}2$ and 250 at each $K\in\{4,6,8,10\}$:
$N_{\mathrm{base}}=1{,}000+4\times250=2{,}000$.
Its top, bottom, random, and no-intervention arms require 7{,}992
replays. At $K{=}10$, 246 trajectories have multiple eligible writes
and four have only one, requiring four and two arms, respectively.
The subsequent matched predecessor-reversion study uses only the 250
$K{=}4$ trajectories. It adds three reversion and three same-anchor
keep runs per trajectory, reusing the 250 no-intervention runs:
\begin{equation}
\begin{aligned}
N_{\mathrm{blank}}&=4{,}000+3\times1{,}000+992\\
                 &=7{,}992,\\
N_{\mathrm{revert+keep}}&=250\times(3+3)=1{,}500,\\
N_{\mathrm{total}}&=7{,}992+1{,}500=9{,}492.
\end{aligned}
\end{equation}
Thus, the two protocols require approximately 9.5k replays in total.
\Cref{tab:behavioral} reports the matched $K{=}4$ study.

\paragraph{Matched predecessor reversion.}
Replacing the selected cumulative memory $m_{i,t}$ with $m_{i,t-1}$
removes its latest increment while retaining the preceding payload.
The top and bottom arms rank non-terminal writes by $\Aoff$; ties
are resolved by step index. The random arm excludes these extremes
in the matched study (zero collisions in 250 trajectories).
Each trajectory contributes all three arms, yielding
$250\times3=750$ trajectory--arm interventions. These are correlated
observations from 250 trajectories.
For a first write without a predecessor, the intervention falls back
to blank memory. The agent resumes greedy decoding with real retrieval.
The keep-adjusted effect is
$d_{i,a}=\mathrm{F1}_{i,a}^{\mathrm{keep}}-
\mathrm{F1}_{i,a}^{\mathrm{revert}}$ for arm $a$.
Arm means average $d_{i,a}$ equally over the 250 trajectories;
95\% CIs resample trajectories with 10{,}000 bootstrap replicates.

\subsection{Results}
\label{app:behavioral_results}

\begingroup
\setlength{\intextsep}{3pt}
\captionsetup{skip=4pt}
\begin{table}[!t]
\centering
\footnotesize
\setlength{\tabcolsep}{4pt}
\renewcommand{\arraystretch}{1.0}
\begin{tabular*}{\columnwidth}{@{\extracolsep{\fill}}lc@{}}
\toprule
\textbf{Reversion arm} & \textbf{F1 drop [95\% CI]} \\
\midrule
Top-ranked by $\Aoff$ & \textbf{0.030} $[0.003,0.057]$ \\
Random write & 0.011 $[-0.018,0.039]$ \\
Bottom-ranked by $\Aoff$ & 0.003 $[-0.024,0.030]$ \\
\midrule
\multicolumn{2}{@{}l}{\textbf{Credit--effect alignment} (750 interventions)} \\
\midrule
\method{} $\Aoff$ & $\rho=0.16$ ($p<10^{-4}$) \\
\mempo{} reference & $\rho=-0.08$ ($p=0.035$) \\
\bottomrule
\end{tabular*}
\caption{Matched predecessor-reversion on the 250-trajectory $K{=}4$
Local Wiki Search subset. All arm estimates include trajectory-bootstrap
95\% CIs. Correlations pair the selected write's offline credit with
its same-anchor keep-adjusted F1 effect.}
\label{tab:behavioral}
\end{table}
\endgroup

The top arm has the largest mean F1 drop and is the only arm whose
CI excludes zero; top $>$ random $>$ bottom also holds for EM.
Excluding first-write fallbacks gives a bottom-arm F1 drop of $-0.004$.
The Spearman correlations use the 750 trajectory--arm observations;
their $p$-values are two-sided asymptotic tests of $H_0:\rho=0$.


\FloatBarrier
\section{Robustness of the Credit Estimator to Target Noise}
\label{app:target_noise}

Both components of \method{}'s memory-specific credit depend on the
target outcome $\zstar$. The difference defining $\Delta_{i,t}$
approximately cancels memory-independent additive shifts, and
hindsight centering removes shifts shared within its normalization pool.
A coherent-but-wrong target can alter both local utility and hindsight
relevance. We test this case alongside paraphrasing and masking.

\subsection{Protocol}
\label{app:target_noise_protocol}

We hold the trained policy fixed and re-score the same 1{,}571 writes
from 250 Local Wiki Search trajectories on the $K{=}4$ split under four
target conditions: (i) an LLM paraphrase, (ii) nominal 30\% contiguous
masking, (iii) nominal 50\% masking, and (iv) a plausible-but-wrong
answer. Masking replaces a contiguous span of whitespace-delimited
tokens with \texttt{[MASKED]}; the span length is the rounded nominal
fraction of the target length, bounded between one and the full length.
Span selection is deterministic given the seed, trajectory, and axis.
The LLM axes use MiniMax-M2.7; template fallbacks occur for 0/250
paraphrases and 4/250 wrong answers and remain included.

We compare recorded credit $\Aoff$ and filter outputs under clean and
perturbed targets, pairing signals by trajectory and write index.
A filter-decision flip is a change in whether the recorded filter weight
exceeds 0.5. Intervention diagnostics use the same cohort and
random stream, with 1{,}000 paired memory-drop events and 451 corrupted
write events per condition. We use 10{,}000 trajectory-clustered
bootstrap replicates for 95\% CIs ($\approx$16k trajectory re-scorings).

Here $\Aoff$ is the smoothed evaluation credit defined in
\Cref{app:behavioral}.

\subsection{Results}
\label{app:target_noise_results}

\begin{table}[!t]
\centering
\footnotesize
\setlength{\tabcolsep}{3pt}
\renewcommand{\arraystretch}{1.15}
\resizebox{\columnwidth}{!}{%
\begin{tabular}{@{}lccc@{}}
\toprule
\textbf{Target condition} &
\makecell{\textbf{Filter-decision}\\\textbf{flip}} &
\makecell{\textbf{Mem-drop}\\\textbf{localization}} &
\makecell{\textbf{Penalty ratio}\\\textbf{\method{}/\mempo{}}} \\
\midrule
Clean                & --      & 0.737 & 0.528 \\
Paraphrase           & 1.78\%  & 0.752 & 0.497 \\
Mask 30\%            & 2.48\%  & 0.721 & 0.460 \\
Mask 50\%            & 2.67\%  & 0.706 & 0.365 \\
Plausible but wrong  & 2.42\%  & 0.734 & 0.491 \\
\bottomrule
\end{tabular}
}
\caption{Controlled target-perturbation study on the matched $K{=}4$
robustness subset (values under \emph{Clean} therefore differ from
\Cref{tab:deconfounding}). Penalty ratios compare pooled mean absolute pre-mask memory credit
with the \mempo{} reference on 451 corrupted write events.}
\label{tab:target_noise}
\end{table}

In \Cref{tab:target_noise}, the retrospective filter's binary decision flips for at most 2.67\%
of the 1{,}571 paired writes. The four flip counts are 28 (paraphrase),
39 (30\% masking), 42 (50\% masking), and 38 (plausible but wrong).

Memory-drop localization changes by $-3.1$ to $+1.5$ points
from the clean condition (0.706--0.752 vs.\ 0.737). The
plausible-but-wrong condition differs by $-0.3$ points. Assessed with
trajectory-clustered paired bootstrap 95\% CIs (10{,}000 replicates,
250 trajectories), the noisy$-$clean difference includes zero on every
axis: paraphrase $+0.015$ $[-0.019, +0.049]$; mask 30\% $-0.016$
$[-0.051, +0.021]$; mask 50\% $-0.031$ $[-0.067, +0.006]$;
plausible-but-wrong $-0.003$ $[-0.039, +0.034]$.

The \method{}/\mempo{} absolute-credit ratio in the
tool-corruption diagnostic remains below 1 under every condition.
The pooled mean absolute pre-mask credit of \method{} decreases
from $0.0154$ with clean targets to $0.0091$ at 50\% masking;
the \mempo{} reference changes from $0.0292$ to $0.0250$.
The methods' trajectory-clustered 95\% CIs do not overlap
in any condition.


\end{document}